\documentclass[10pt,twocolumn,letterpaper]{article}

\usepackage{iccv}
\usepackage{times}
\usepackage{epsfig}
\usepackage{graphicx}
\usepackage{amsmath}
\usepackage{amssymb}
\usepackage{paralist}
\usepackage{booktabs}
\usepackage{bbm}

\usepackage[pagebackref=true,breaklinks=true,letterpaper=true,colorlinks,bookmarks=false]{hyperref}

\iccvfinalcopy % *** Uncomment this line for the final submission

\newcommand*{\resp}{resp.\@\xspace}

\DeclareMathOperator*{\argmax}{\arg\!\max}

\newcommand{\est}[1]{\hat{#1}}
\newcommand{\gt}[1]{{#1}^*}

\newcommand{\crd}{\mathbf{y}}
\newcommand{\crds}{\mathcal{Y}}

\newcommand{\mdl}{\mathbf{h}}
\newcommand{\pool}{\mathcal{H}}
\newcommand{\loss}{\ell}
\newcommand{\Loss}{\mathcal{L}}

\newcommand{\param}{\mathbf{w}}

\newcommand{\expectation}[2]{\mathbb{E}_{#1}\left[ #2 \right]}
\newcommand{\derv}[1]{\frac{\partial}{\partial #1}}

\begin{document}

%%%%%%%%% TITLE
\title{Expert Sample Consensus Applied to Camera Re-Localization}

\author{Eric Brachmann and Carsten Rother\\
Visual Learning Lab\\
Heidelberg University (HCI/IWR)\\
{\tt\small http://vislearn.de}
}

\maketitle

%%%%%%%%% ABSTRACT
\begin{abstract}

Fitting model parameters to a set of noisy data points is a common problem in computer vision.
In this work, we fit the 6D camera pose to a set of noisy correspondences between the 2D input image and a known 3D environment. 
We estimate these correspondences from the image using a neural network. 
Since the correspondences often contain outliers, we utilize a robust estimator such as Random Sample Consensus (RANSAC) or \emph{Differentiable RANSAC} (DSAC) to fit the pose parameters.
When the problem domain, \eg the space of all 2D-3D correspondences, is large or ambiguous, a single network does not cover the domain well. 
\emph{Mixture of Experts (MoE)} is a popular strategy to divide a problem domain among an ensemble of specialized networks, so called experts, where a gating network decides which expert is responsible for a given input.
In this work, we introduce \emph{Expert Sample Consensus} (ESAC), which integrates DSAC in a MoE. 
Our main technical contribution is an efficient method to train ESAC jointly and end-to-end. 
We demonstrate experimentally that ESAC handles two real-world problems better than competing methods, \ie scalability and ambiguity.
We apply ESAC to fitting simple geometric models to synthetic images, and to camera re-localization for difficult, real datasets.   
   
\end{abstract}

%%%%%%%%% BODY TEXT
\vspace{-0.3cm}
\section{Introduction}
\vspace{-0.1cm}

\begin{figure}[t]
\begin{center}
   \includegraphics[width=\linewidth]{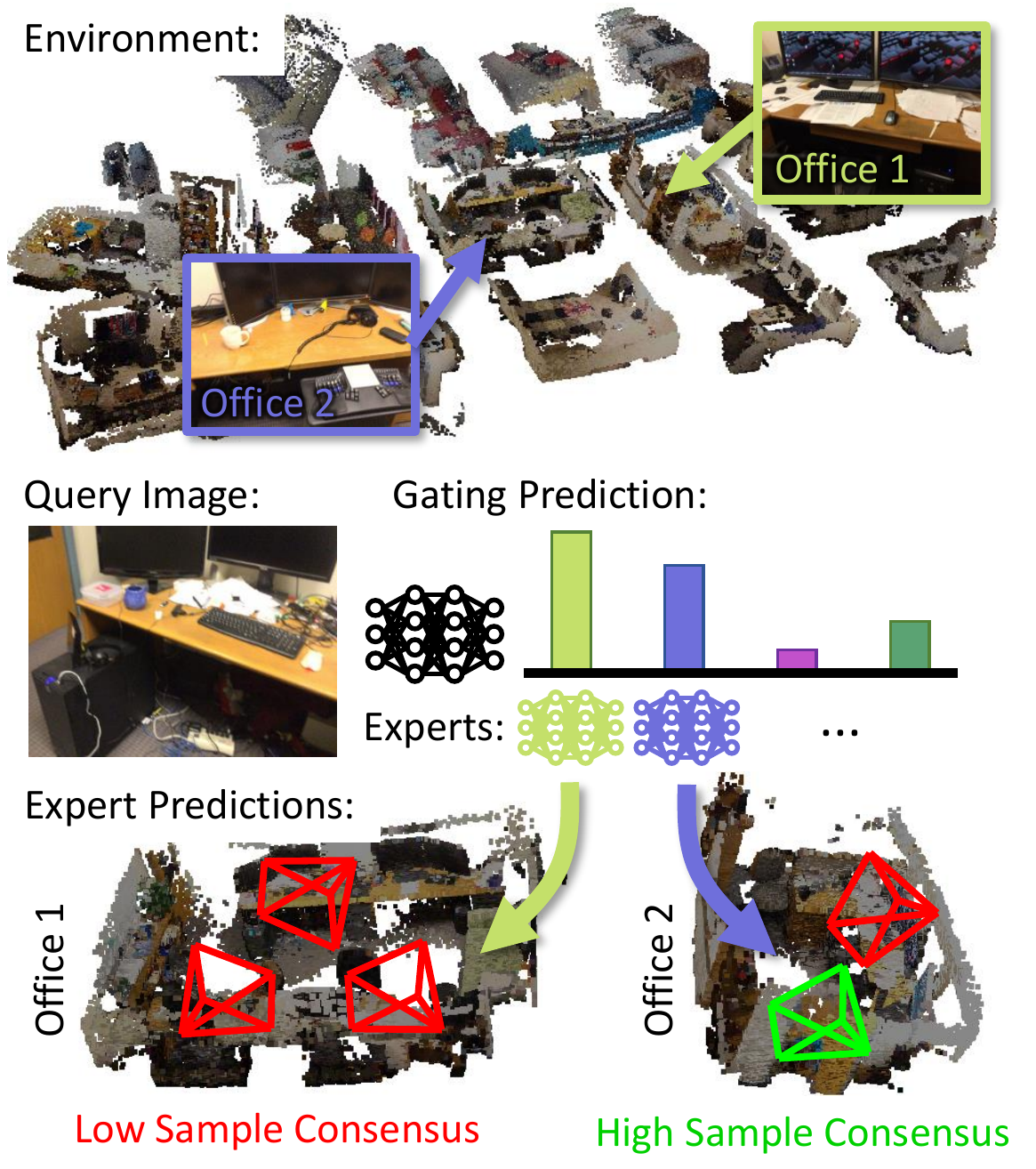}
\end{center}
   \vspace{-0.4cm}
   \caption{\textbf{Camera Re-Localization Using ESAC.} Given an environment consisting of several ambiguous rooms (top) and a query image (middle), we estimate the 6D camera pose (bottom). A gating network (black) predicts a probability for each room. We distribute a budget of pose hypotheses to expert networks specialized to each room. We choose the pose hypothesis with maximum sample consensus (green), \ie the maximum geometric consistency. We train all networks jointly and \mbox{end-to-end}.}
   \vspace{-0.4cm}
\label{fig:teaser}
\end{figure}

In computer vision, we often have a model that explains an observation with a small set of parameters. 
For example, our model is the 6D pose (translation and rotation) of a camera, and our observations are images of a known 3D environment. 
The task of camera re-localization is then to robustly and accurately predict the 6D camera pose given the camera image.
However, inferring model parameters from an observation is difficult because many effects are not explained by our model.
People might move through the environment, and its appearance varies largely due to lighting effects such as day versus night.
We usually map our observation to a representation from which we can infer model parameters more easily.
For example, in camera re-localization we can train a neural network to predict correspondences between the 2D input image and the 3D environment.
Inferring the camera pose from these correspondences is much easier, and various geometric solvers for this problem exist \cite{kabsch1976solution, gao2003complete, lepetit2009epnp}.
Because some predictions of the network might be erroneous, \ie we have outlier correspondences, we utilize a robust estimator such as \emph{Random Sample Consensus} (RANSAC) \cite{ransac1981}, \resp its differentiable counterpart \emph{Differentiable Sample Consensus} (DSAC) \cite{brachmann2017dsac}, or other differentiable estimators \cite{goodcorr18, deepfund18} for training.

For some tasks, the problem domain is large or ambiguous. 
In camera re-localization, an environment could feature repeating structures that are unique {\it locally} but not globally, \eg office equipment, radiators or windows.
A single feed-forward network cannot predict a correct correspondence for such objects because there are multiple valid solutions.
However, if we train an ensemble of networks where each network specializes in a local part of the environment, we can resolve such ambiguities.
This strategy is known in machine learning as \emph{Mixture of Experts (MoE)} \cite{Jacobs1991}.
Each expert is a network specialized to one part of the problem domain. 
An additional gating network decides which expert is responsible for a given observation.
More specifically, the output of the gating network is a categorical distribution over experts, which either guides the selection of a single expert, or a weighted average of all expert outputs \cite{Masoudnia2014}. 

In this work, we extend Mixture of Experts for fitting parametric models.
Each expert specializes to a part of all training observations, and predicts a representation to which we fit model parameters using DSAC.
We argue that two realizations of a Mixture of Experts model are not optimal: i) letting the gating network select one expert only \cite{netdis15, hdcnn15, Aljundi2017ExpertGL, moenn17}; ii) giving as output a weighted average of all experts \cite{Jacobs1991, noe16}.  
In the first case, we ignore that the gating network might attribute substantial probability to more than one expert.
We might choose the wrong expert, and get a poor result.
In the second case, we calculate an average in model parameter space which can be instable in learning \cite{brachmann2017dsac}. 
In our realization of a Mixture of Experts model, we integrate the gating network into the hypothesize-and-verify framework of DSAC. 
To estimate model parameters, DSAC creates many model hypotheses by sampling small subsets of data points, and fitting model parameters to each subset.
DSAC scores hypotheses according to their consistency with all data points, \ie their sample consensus.
One hypothesis is selected as the final estimate according to this score.
Hypothesis selection is probabilistic, and training aims at minimizing the expected task loss.

Instead of letting the gating network pick one expert, and fit model parameters only to this expert's prediction, we distribute model hypotheses among experts.
Each expert receives a share of the total number of hypotheses according to the gating network.
For the final selection, we score each hypothesis according to sample consensus, irrespective of what expert it came from, see Fig~\ref{fig:teaser}.
Therefore, as long as the gating network attributes some probability to the correct expert, we can still get an accurate model parameter estimate.
We call this framework \emph{Expert Sample Consensus} (ESAC). 
We train the network ensemble jointly and end-to-end by minimizing the expected task loss.
We define the expectation over both, hypotheses sharing according to the gating network, and hypothesis selection according to sample consensus.

We demonstrate our method on a toy problem where the gating network has to decide which model to fit to synthetic data - a line or a circle. 
Compared to naive expert selection, our method proves to be extremely robust regarding the gating network's ability to assign the correct expert. 
Our method also achieves state-of-the-art results in camera re-localization where each expert specializes in a separate, small part of a larger indoor environment.

\noindent We give the following main \emph{contributions}:
\begin{compactitem}
\item We present \emph{Expert Sample Consensus} (ESAC), an ensemble formulation of Differentiable Sample Consensus (DSAC) which we derive from Mixture of Experts (MoE).

\item A method to train ESAC jointly and end-to-end.

\item We demonstrate the properties of our algorithm on a toy problem of fitting simple parametric models to noisy, synthetic inputs.

\item Our formulation improves on two real-world aspects of learning-based camera re-localization, scalability and ambiguity. We achieve state-of-the-art results on difficult, public datasets for indoor re-localization.
\end{compactitem}

\section{Related Work}
\label{sec:related}

\noindent{\bf Ensemble Methods.}
To improve the accuracy of machine learning algorithms, one can train multiple base-learners and combine their predictions.
A common strategy is averaging, so that errors of individual learners cancel out \cite{Breiman2001, alexnet2012, Simonyan2014, resnet2015}.
To ensure that base-learners produce non-identical predictions, they are trained using random subsets of training data (bagging) or using random initializations of parameters (\eg network weights).
Boosting refers to a weighted average of predictions where the weights emerge from each base-learners ability to classify training samples \cite{Freund99ashort}.
In these ensemble methods, all base-learners are trained on the full problem domain.

In contrast, Mixture of Experts (MoE) \cite{Jacobs1991} employs a divide-and-conquer strategy where each base-learner, \resp expert, specializes in one part of the problem domain. 
An additional gating network assesses the relevancy of each expert for a given input, and predicts an associated weight.
The ensemble prediction is a weighted average of the experts' outputs.
MoE has been trained by minimizing the expected training loss \cite{Jacobs1991}, maximizing the likelihood under a Gaussian mixture model interpretation \cite{Jacobs1991} or using the expectation-maximization (EM) algorithm \cite{hiermoe}.

MoE has been applied to image classification where each expert specializes to a subset of classes \cite{hdcnn15, netdis15, noe16, Aljundi2017ExpertGL}.
Ahmed \etal \cite{noe16} find disjunct subsets by an EM-style algorithm. 
Hinton \etal \cite{netdis15} and Yan \etal \cite{hdcnn15} find subsets of classes based on class confusion of a generalist base network. 
Aljundi \etal \cite{Aljundi2017ExpertGL} apply MoE to lifelong multi-task learning.
Whenever their system should be extended with a new task (\eg a new object class) they train a new expert and a new expert gate. 
Each expert gate measures the similarity of an input with its associated task, and the gate with the highest similarity forwards the input to its expert.

In all aforementioned methods, the experts' outputs constitute the ensemble output directly. 
In contrast, we are interested in a scenario where experts output a representation to which we fit parametric models in a robust fashion while maintaining the ability to train the ensemble jointly and end-to-end.
To the best of our knowledge, this has not been addressed, previously.
Some of the aforementioned methods make use of conditional computation, \ie the gating network selects a subset of experts to evaluate while others stay idle \cite{hdcnn15, netdis15, Aljundi2017ExpertGL}.
While this is computationally efficient, routing errors can occur, \ie selection of the incorrect expert results in catastrophic errors.
In this work, we distribute computational budget between experts based on the potentially soft prediction of the gating network.
Thereby, we strike a good balance between efficiency and robustness.

\noindent{\bf Camera Re-Localization.}
Camera re-localization has been addressed with a very diverse set of methods.
Some authors use image-based retrieval systems \cite{schindler2007city, cao2013graph, netvlad2016} to map a query image to the nearest neighbor in a set of database images with known pose. 
Pose regression methods \cite{kendall2015convolutional, LSTMPoseNet, geometricloss, relocnet2018, mapnet2018} train neural feed-forward networks to predict the 6D pose directly from an input image.
Pose-regression methods vary in network architecture, pose parametrization, or training loss.
Both, retrieval-based and pose-regression methods, are very efficient but limited in accuracy.
Feature-based re-localization methods \cite{lim2012real, sattler2015hyperpoints, sattler2016efficient, sattler2016large, sattler2017largescale, Toft2018ECCV} match sparse feature points of the input image to a sparse 3D reconstruction of the environment.
The 6D camera pose is estimated from these 2D-3D correspondences using RANSAC.
These methods are very accurate, scale well but have problems with texture-less surfaces and image conditions like motion blur because the feature detectors fail \cite{shotton13scorf, kendall2015convolutional}. 

Scene coordinate regression methods \cite{shotton13scorf, guzman2014multi, valentin2015cvpr, brachmann2016, rfvscnn2016, meng17, brachmann2017dsac, cavallari2017fly, meng18, brachmann2018lessmore} also estimate 2D-3D correspondences between image and environment but do so densely for each pixel of the input image.
This circumvents the need for a feature detector with the aforementioned draw-backs of feature-based methods.
Brachmann \etal \cite{brachmann2017dsac} combine a neural network for scene coordinate regression with a differentiable RANSAC for an end-to-end trainable camera re-localization pipeline.
Brachmann and Rother \cite{brachmann2018lessmore} improve the pipeline's initialization and differentiable pose optimization to achieve state-of-the-art results for indoor camera re-localization from single RGB images.
We build on and extend \cite{brachmann2017dsac, brachmann2018lessmore} by combining them with our ESAC framework. 
Thereby, we are able to address two real-world problems: scalability and ambiguity in camera re-localization. 
Some scene coordinate regression methods use an ensemble of base learners, namely random forests \cite{shotton13scorf, valentin2015cvpr, brachmann2016, rfvscnn2016, meng17, cavallari2017fly, meng18}.
Guzman-Rivera \etal \cite{guzman2014multi} train the random forest in a boosting-like manner to diversify its predictions.
Massiceti \etal \cite{rfvscnn2016} map an ensemble of decision trees to an ensemble of neural networks.
However, in none of these methods do the base-learners specialize in parts of the problem domain.

In \cite{brachmann2016}, Brachmann \etal train a joint classification-regression forest for camera re-localization.
The forest classifies which part of the environment an input belongs to, and regresses relative scene coordinates for this part.
More recently, image-retrieval and relative pose regression have been combined in one system for good accuracy in \cite{taira2018inloc}.
Both works, \cite{brachmann2016} and \cite{taira2018inloc}, bear some resemblance to our strategy but utilize one large model without the benefit of efficient, conditional computation.
Also, their models cannot be trained in an end-to-end fashion. 

\noindent{\bf Model Selection.} Sometimes, the model type has to be estimated concurrently with the model parameters.
E.g.~data points could be explained by a line or higher order polynomials.
Methods for model selection implement a trade-off between model expressiveness and fitting error \cite{akaike74, schwarz78}.
For illustrative purposes, we introduce ESAC on a toy problem where it learns model selection in a supervised fashion.
However, in our main application, camera re-localization, the model type is always known to be a 6D pose.

\section{Method}
\label{sec:method}
We start by reviewing DSAC \cite{brachmann2017dsac} for fitting parametric models in Sec.~\ref{sec:method:dsac}.
Then, in Sec.~\ref{sec:method:moe}, we introduce Mixture of Experts \cite{Jacobs1991} with expert selection.
Finally, we present ESAC, an ensemble formulation of DSAC in Sec.~\ref{sec:method:esac}.
We will explain these concepts for a simple toy problem before applying them to camera re-localization in Sec.~\ref{sec:camloc}.

\subsection{Differentiable Sample Consensus}
\label{sec:method:dsac}

\begin{figure}[t]
\begin{center}
\vspace{-0.3cm}
   \includegraphics[width=0.95\linewidth]{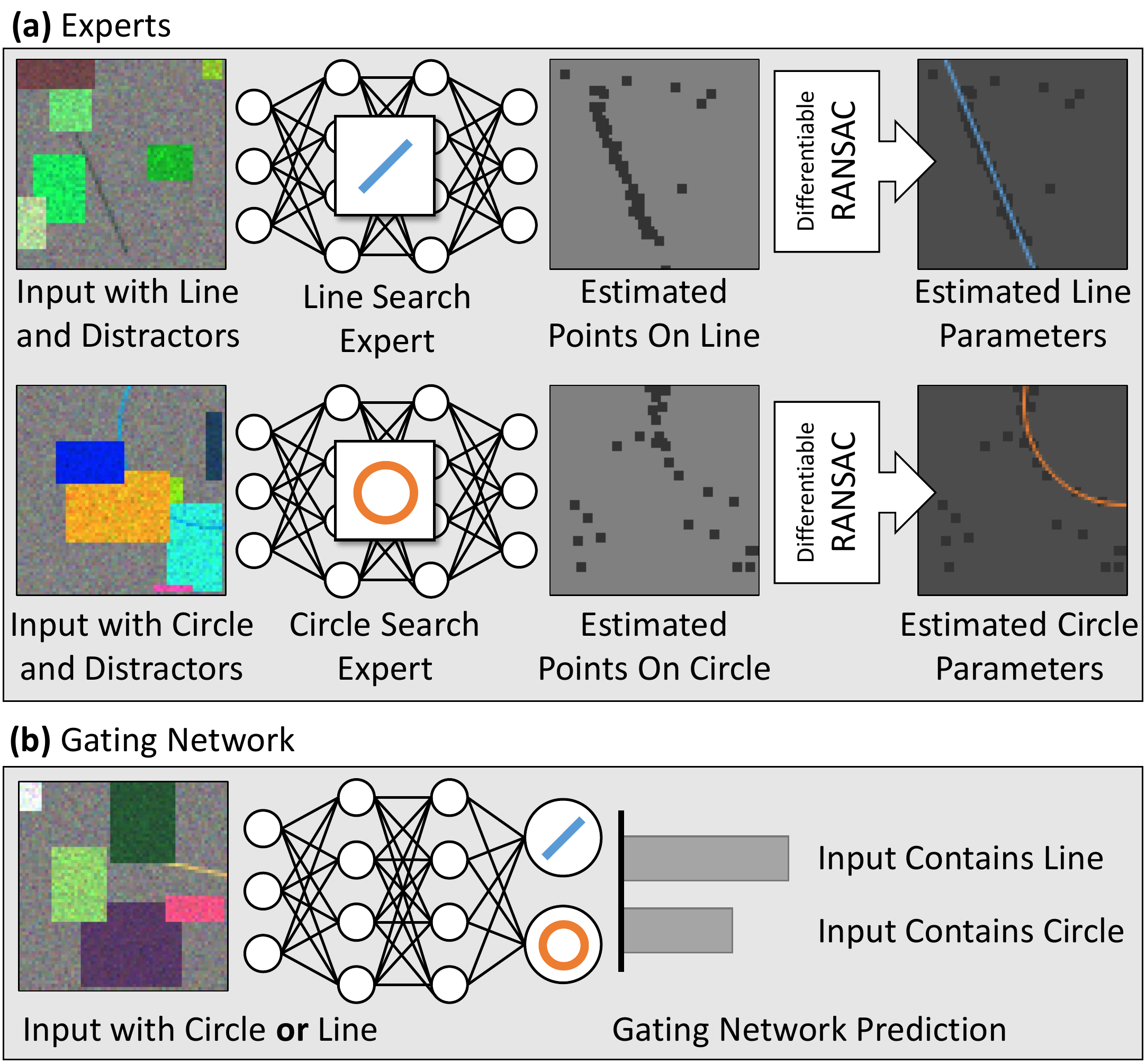}
\end{center}
\vspace{-0.3cm}
   \caption{\textbf{Network Ensemble for a Toy Problem. a)}    
   Two expert networks, one specialized to finding lines, one specialized to finding circles. 
   Both experts predict a set of 2D points which should lie on the line or circle, respectively.
   We fit model parameters to these points using differentiable RANSAC.
   \textbf{b)} The gating network predicts whether an image contains either a line or a circle.}
\label{fig:method:ensemble}
\vspace{-0.3cm}
\end{figure}

We are interested in estimating a set of model \mbox{parameters $\mdl$} given an observation $I$.
For instance, the model could be a 2D line with slope $m$ and intercept $n$, \ie $\mdl = (m, n)$.
Observation $I$ is an image of the line which also contains noise and distractors which are not explained by our model $\mdl$.
See top of Fig.~\ref{fig:method:ensemble} a) for an example input $I$ where the distractors are boxes that partly occlude the line.

Instead of fitting model parameters $\mdl$ directly to $I$, we deduce an intermediate representation $\crds$ from $I$ to which we can fit our model easily.
In the case of a line, $\crds$ could be a set of 2D points $\crd \in \crds$ with $\crd=(y_0, y_1)$, where each point is explained by our model: $y_1 = m y_0 + n$.
We can deduce line parameters $\mdl$ from $\crds$ using linear regression or Deming regression \cite{deming1943statistical}.

Since the image formation process is complicated and/or unknown to us, there is no simple way to infer $\crds$ from $I$.
Instead, we train a neural network $f$ with learnable parameters $\param$ to predict $\crds = f(I; \param)$.
The neural network can learn to ignore distractors and image noise to some extent. 
However, it is likely to make some mistakes, \eg predict some points $\crd$ not explained by our model $\mdl$.
Therefore, we employ a robust estimator $\est{\mdl}$, namely Random Sample Consensus (RANSAC) \cite{ransac1981}, and, for neural network training, Differentiable Sample Consensus (DSAC) \cite{brachmann2017dsac}.

\noindent{\bf RANSAC.}
RANSAC robustly estimates model parameters by sampling a pool of $N$ model hypotheses $\mdl_j$ with \mbox{$j \in \{1, \dots, N\}$}.
A hypothesis is sampled by randomly choosing a minimal set from $\crds$ and fitting model parameters to it.
For a 2D line, a minimal set consists of two 2D points which determine slope and intercept.
Each hypothesis is scored by measuring its sample consensus or inlier count $s(\cdot)$, \ie the number of data points $\crd$ that agree with the hypothesis.
\begin{equation}
\label{eq:ransac:in}
s(\mdl,\crds) = \sum_{\crd \in \crds} \mathbbm{1}(\tau - d(\crd, \mdl)),
\end{equation}
where $d(\crd, \mdl)$ is a measure of distance between model hypothesis $\mdl$ and data point $\crd$, \eg the point-line distance.
Parameter $\tau$ is a threshold that encapsulates our tolerance for inlier errors, and $\mathbbm{1}(\cdot)$ denotes the Heaviside step function.
Our final estimate is the model hypothesis with the maximum score: 
\begin{equation}
\label{eq:ransac}
\est{\mdl} = \mdl_j ~ \text{with} ~ j = \argmax_j s(\mdl_j, \crds)
\end{equation}
Due to the non-differentiability of the $\argmax$ selection, we cannot use RANSAC directly in neural network training.
However, Brachmann \etal \cite{brachmann2017dsac} proposed a differentiable version of the algorithm which we will discuss next.

\noindent{\bf DSAC.}
The core idea of Differentiable Sample Consensus \cite{brachmann2017dsac} is to make hypothesis selection probabilistic.
Instead of choosing the hypothesis with maximum score deterministically as in Eq.~\ref{eq:ransac}, we choose it randomly according to a softmax distribution over scores:
\begin{equation}
\label{eq:dsac:forward}
\est{\mdl} = \mdl_j ~ \text{with} ~  j \sim p(j) = \frac{\text{exp}(s(\mdl_j,\crds))}{\sum_{j'}\text{exp}(s(\mdl_{j'},\crds))}
\end{equation}
This allows us to minimize the expected task loss $\Loss(\param)$ during training:
\begin{equation}
\label{eq:dsac}
\Loss(\param) = \expectation{j \sim p(j)}{\loss(\mdl_j)},
\end{equation}
where $\loss(\mdl)$ measures the error of a model hypothesis $\mdl$ \wrt some ground truth parameters $\gt{\mdl}$.
Since $\Loss(\param)$ is a weighted sum with a finite number of $N$ summands, one for each hypothesis in our pool, we can calculate it and its gradients exactly.
As one last consideration, we have to replace the non-differentiable inlier count of Eq.~\ref{eq:ransac:in} by a soft version \cite{brachmann2018lessmore}.
\begin{equation}
\label{eq:dsac:in}
s(\mdl,\crds) = \alpha \sum_{\crd \in \crds} 1 - \text{sig}(\beta d(\crd, \mdl)- \beta \tau),
\end{equation}
where $\text{sig}(\cdot)$ denotes the Sigmoid function, and $\alpha, \beta$ are hyperparameters which control the softness of the score \cite{brachmann2018lessmore}.

By minimizing $\Loss(\param)$, we can train our network $f(I;\param)$ in an end-to-end fashion using DSAC. 
The network learns to predict a representation $\crds$ that yields an accurate model estimate $\est{\mdl}$, although $\crds$ might still contain outliers.
For the toy problem of fitting a 2D line, we show an example run of the full pipeline in Fig.~\ref{fig:method:ensemble} a) top.

\subsection{Expert Selection}
\label{sec:method:moe}

In the following, we introduce the notion of experts for the scenario of parametric model fitting. 
Firstly, we apply the original formulation of Mixture of Experts (MoE) \cite{Jacobs1991} before extending it in Sec.~\ref{sec:method:esac}.

Instead of training one neural network responsible for all inputs, we train an ensemble of $M$ experts $f_e(I; \param)$ with $e \in \{1, \dots, M \}$.
We denote the output of each expert with $\crds_e$.
A gating network $g(e, I; \param)$ decides for a given input $I$ which expert is responsible, \ie it predicts a probability distribution over experts: $p(e) = g(e, I; \param)$.
For notation simplicity we stack the learnable parameters of all individual networks in a single parameter vector $\param$.

For illustration, we change the toy problem of the previous section in the following way.
Some inputs $I$ show a 2D line (as before) while others show a 2D circle.
Therefore, we extend our model parameters to $\mdl = (m, n, r)$.
In case of a circle, $(m, n)$ is the circle center and $r$ is its radius.
In case of a line, $m$ and $n$ are slope and intercept, respectively and we set $r = -1$ to indicate it is not a circle.

We train two experts, \eg $M=2$, one specialized for fitting lines, one specialized for fitting circles.
Additionally, we train a gating network which should decide for an arbitrary input whether it shows a line or a circle, so that we can apply the correct expert.
See Fig.~\ref{fig:method:ensemble} for a visualization of all three networks and their respective task.

\begin{figure*}
\begin{center}
\includegraphics[width=0.9\linewidth]{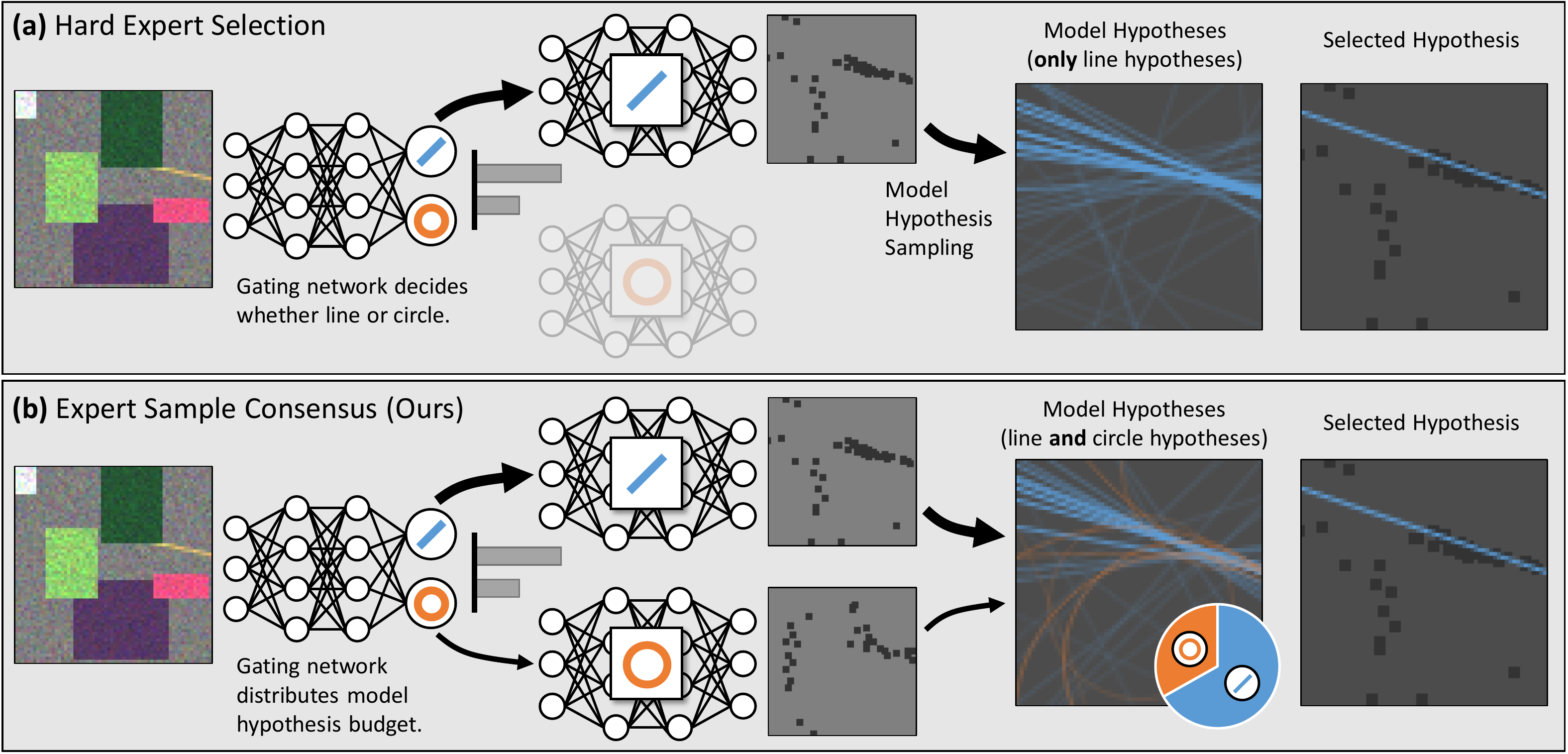}
\end{center}
\vspace{-0.4cm}
   \caption{\textbf{Ensemble Interplay.} Given an image of a line or a circle, we estimate the parameters of the associated model. \textbf{a)}
    The gating network chooses one expert for a given input.
    We sample model hypotheses only based on this expert's prediction.
    \textbf{b)} The gating network predicts how the number of model hypotheses should be divided among experts, \ie we sample line \textbf{and} circle hypotheses. 
    In this example, the estimate of \textbf{a)} and \textbf{b)} is similar, but in \textbf{b)} we incorporate the full prediction of the gating network, instead of only the largest probability.}
\label{fig:method:system}
\vspace{-0.4cm}
\end{figure*}

Given an image $I$, we first choose an expert according to the gating network prediction $e \sim p(e)$.
We let this expert estimate $\crds_e$, and apply DSAC, \ie we sample a pool of hypotheses from $\crds_e$.
We choose our estimate similar to Eq.~\ref{eq:dsac:forward} according to
\begin{equation}
\label{eq:moe:forward}
\est{\mdl} = \mdl_j~ \text{with} ~  j \sim p(j|e) = \frac{\text{exp}(s(\mdl_j,\crds_e))}{\sum_{j'}\text{exp}(s(\mdl_{j'},\crds_e))}.
\end{equation}
We illustrate the forward process of the ensemble in \mbox{Fig.~\ref{fig:method:system} a)}.
To train the network ensemble, we can adapt the training formulation of DSAC (Eq.~\ref{eq:dsac}) in the following way.
\begin{equation}
\label{eq:moe}
\Loss(\param) = \mathbb{E}_{e \sim p(e)}\expectation{j \sim p(j|e)}{\loss(\mdl_j)},
\end{equation}
\ie we minimize the expected loss over choosing the correct expert according to $p(e)$, and selecting a model hypothesis from this expert according to $p(j|e)$.
Note, that we enforce specialization of experts in this training formulation by running the appropriate version of DSAC depending on which expert we chose, \ie we fit either a circle or a line to $\crds_e$.

To calculate the outer expectation, we have to sum over all $M$ experts and run DSAC each time for the inner expectation. 
Since DSAC is costly, and in some applications we might have a large number of experts, this can be infeasible.
However, we can re-write the gradients of the expectation as an expectation itself \cite{brachmann2017dsac}.
This allows us to efficiently approximate the gradients via sampling.
\begin{gather}
%\derv{\param} \Loss(\param) = \derv{\param} \mathbb{E}_{\pool \sim p(\pool)}\expectation{j \sim p(j|\pool)}{\loss(\mdl_j)} \nonumber \\
%\derv{\param} \Loss(\param) = \nonumber \\
% \expectation{e}{\expectation{j}{\loss(\mdl_j)} \derv{\param} \log p(e) + \derv{\param} \expectation{j}{\loss(\mdl_j)}}
\derv{\param} \Loss(\param) = \expectation{e}{\expectation{j}{\loss} \derv{\param} \log p(e) + \derv{\param} \expectation{j}{\loss}} \nonumber \\
\approx \frac{1}{K} \sum_{k=1}^K \left[ \expectation{j}{\loss} \derv{\param} \log p(e_k) + \derv{\param} \expectation{j}{\loss} \right],
\end{gather}
where we sample $e_k \sim p(e)$ $K$ times and average the gradients. 
We use the abbreviations $\mathbb{E}_{e}$, $\mathbb{E}_{j}$ and $\loss$ for the respective entities in Eq.~\ref{eq:moe}.
In practice, when training with stochastic gradient descent, we can approximate the expectation with $K=1$ sample which means that we do one run of DSAC per training input.

Since we select only one expert at test time, we only have to compute this expert's forward pass, which is computationally efficient. 
However, if we chose the wrong expert, \ie an expert not specialized to current input $I$, we cannot hope to get a sensible prediction $\est{\mdl}$.
Therefore, the accuracy of this MoE formulation is limited by the accuracy of the gating network.
In the next section, we describe our alternative, new formulation which is more robust to inaccuracies of the gating network.

\subsection{Expert Sample Consensus}
\label{sec:method:esac}

Instead of having the gating network select one expert with the risk of selecting the wrong one, we distribute our budget of $N$ model hypotheses among experts.
We sample $n_e \leq N$ hypotheses from each expert's prediction $\crds_e$. 
For this purpose, we define a vector $\pool$ that expresses how many hypotheses we assign to each expert.
\begin{equation}
\pool = (n_1, \dots, n_e, \dots, n_M) ~ \text{with} ~ \sum n_e = N
\end{equation}
We choose $\pool$ for a given input $I$ based on the output of the gating network.
More specifically, $\pool$ follows a multinomial distribution based on the gating probabilities $g(e, I; \param)$.
\begin{equation}
p(\pool) = \frac{N!}{\prod_e n_e!} \prod_e g(e, I; \param)^{n_e}
\end{equation}
Given an image $I$, we first choose $\pool \sim p(\pool)$, and then, according to $\pool$ we sample $n_e$ hypotheses $\mdl_{(e,j)}$ with \mbox{$j \in \{1,\dots, n_e \}$} from each expert prediction $\crds_e$.
We use an index pair $(e,j)$ to denote which expert a hypothesis belongs to, and which of the $n_e$ hypotheses of this expert it is, specifically.
We choose our estimate similar to Eq.~\ref{eq:dsac:forward} and Eq.~\ref{eq:moe:forward} according to
\begin{gather}
\est{\mdl} = \mdl_{(e,j)}~ \text{with} ~  {(e,j)} \sim p(e,j|\pool), ~ \text{and} \nonumber \\
p(e,j|\pool) = \frac{\text{exp}(s(\mdl_{(e,j)},\crds_e))}{\sum_{e'}\sum_{j'}\text{exp}(s(\mdl_{(e',j')},\crds_{e'}))}
\end{gather}
Note that $p(e,j|\pool)$ is a softmax distribution over all $N$ hypotheses, \ie we choose a hypothesis solely based on its score $s(\cdot)$ irrespective of which expert it came from.
In particular, the gating network does not influence hypothesis selection directly, but only guides hypotheses distribution among experts.
Depending on the prediction of the gating network $g(e, I; \param)$, some experts with low probability will have no hypotheses assigned ($n_e = 0$). 
For these experts, we do not need $\crds_e$, and hence can save computing the associated forward pass, implementing conditional computation.
We visualize our method in Fig.~\ref{fig:method:system} b).

For training, we adapt our MoE training objective of Eq.~\ref{eq:moe} and minimize
\begin{equation}
\label{eq:esac}
\Loss(\param) = \mathbb{E}_{\pool \sim p(\pool)}\expectation{(e, j) \sim p(e,j|\pool)}{\loss(\mdl_{(e,j)})}.
\end{equation}
\ie we minimize the expected loss over distributing $N$ hypotheses, and selecting a final estimate.
Since $p(\pool)$ is a distribution over all possible vectors $\pool$, we again rewrite the gradients of $\Loss(\param)$ as an expectation, and approximate via sampling:
\begin{gather}
\label{eq:esac:gradients}
\derv{\param} \Loss(\param)\approx \frac{1}{K} \sum_{k=1}^K \left[ \expectation{e,j}{\loss} \derv{\param} \log p(\pool_k) + \derv{\param} \expectation{e,j}{\loss} \right]
\end{gather}
In practice we found $K=1$ to suffice.
Throughout training, we sample many different hypotheses splits. 
Whenever a responsible expert receives too few hypotheses, Eq.~\ref{eq:esac} yields a large loss, and hence a large training signal for the gating network.
On the other hand, receiving too many hypotheses will not decrease the loss further, and there will be no training signal to reward it.
Therefore, the gating network learns the trade-off between assigning broad distributions $p(e)$ in ambiguous cases, and assigning sufficiently many hypotheses to the most likely experts. 

Calculating the approximate gradients of Eq.~\ref{eq:esac:gradients} involves the derivative of the log probability for a given $\pool$ which we calculate as
\begin{gather}
\derv{\param} \log p(\pool) = \sum_e \frac{n_e}{g(e,I;\param)} \derv{\param} g(e, I;\param).
\end{gather}

\section{ESAC for Camera Re-Localization}
\label{sec:camloc}

We estimate the 6D camera pose $\mdl = (\mathbf{t}, \boldsymbol{\theta})$, consisting of 3D translation $\mathbf{t}$ and 3D rotation $\boldsymbol{\theta}$, from a single RGB image.
Our pipeline is based on DSAC++ of Brachmann and Rother \cite{brachmann2018lessmore} which itself is based on the scene coordinate regression method of Shotton \etal \cite{shotton13scorf}.
For each pixel $i$ with 2D position $\mathbf{p}_i$ in an image, we regress a 3D scene coordinate $\crd_i$, \ie the coordinate of the pixel in world space.

Given a minimal set of four 2D-3D correspondences $(\mathbf{p}_i, \crd_i)$ we can estimate $\mdl$ using a perspective-n-point algorithm \cite{gao2003complete,lepetit2009epnp}.
We employ a robust estimator $\est{\mdl}$ as described in Sec.~\ref{sec:method}.
That is, we sample multiple minimal sets to create a pool of $N$ pose hypotheses $\mdl_j$, and select the best one according to a scoring function. 
We follow \cite{brachmann2018lessmore}, and use a soft inlier count as score. 
See also Eq.~\ref{eq:dsac:in} where we use the re-projection error of a scene coordinate for $d(\crd, \mdl)$.

Once we have chosen a hypothesis, we refine it using the differentiable pose optimization of \cite{brachmann2018lessmore}.
Refinement iteratively resolves the perspective-n-point problem on all inliers of a hypothesis.
Gradients are approximated via a linearizion of the objective function in the last refinement iteration.
Our output is the refined, selected hypothesis $R(\est{\mdl})$.
As task loss for training, we use \mbox{$\loss(\mdl) = \angle(\boldsymbol{\theta}, \gt{\boldsymbol{\theta}}) + \gamma||\mathbf{t}-\gt{\mathbf{t}}||$}, where $\angle(\cdot)$ denotes angle difference. 
The hyperparameter $\gamma$ controls the trade-off between rotation and translation errors \cite{kendall2015convolutional}.
We use $\gamma=100$ when measuring angles in degree and translation in meters.

We estimate scene coordinates $\crd$ using an ensemble of experts $f_e(I;\param)$ and a gating network $g(e, I; \param)$.
When designing the expert network architecture we were inspired by DSAC++ \cite{brachmann2018lessmore}.
Each expert is an FCN \cite{fcn2015} which predicts $80 \times 60$ scene coordinates for a $640 \times 480$px image.
Different from DSAC++ \cite{brachmann2018lessmore}, we use a ResNet architecture \cite{resnet2015} instead of VGG \cite{Simonyan2014}.
We found ResNet to achieve similar accuracy while being more efficient in computation time and memory (28 vs.~210MB).
Each expert has 16 layers, 6M parameters and a $81$px receptive field.
The gating network has 10 layers and 100k parameters. 
The receptive field of the gating network is the complete image, \ie it incorporates more context when assigning experts.
Experts have a small receptive field to be robust to view point changes. 
Our implementation is based on PyTorch \cite{paszke2017automatic}, and we will make it publicly available\footnote{vislearn.de/research/scene-understanding/pose-estimation/\#ICCV19} .

\section{Experiments}
\label{sec:experiments}

We evaluate ESAC for the toy problem introduced in Sec.~\ref{sec:method}, and camera re-localization from single RGB images.

\subsection{Toy Problem}

\begin{figure}
\begin{center}
\includegraphics[width=1\linewidth]{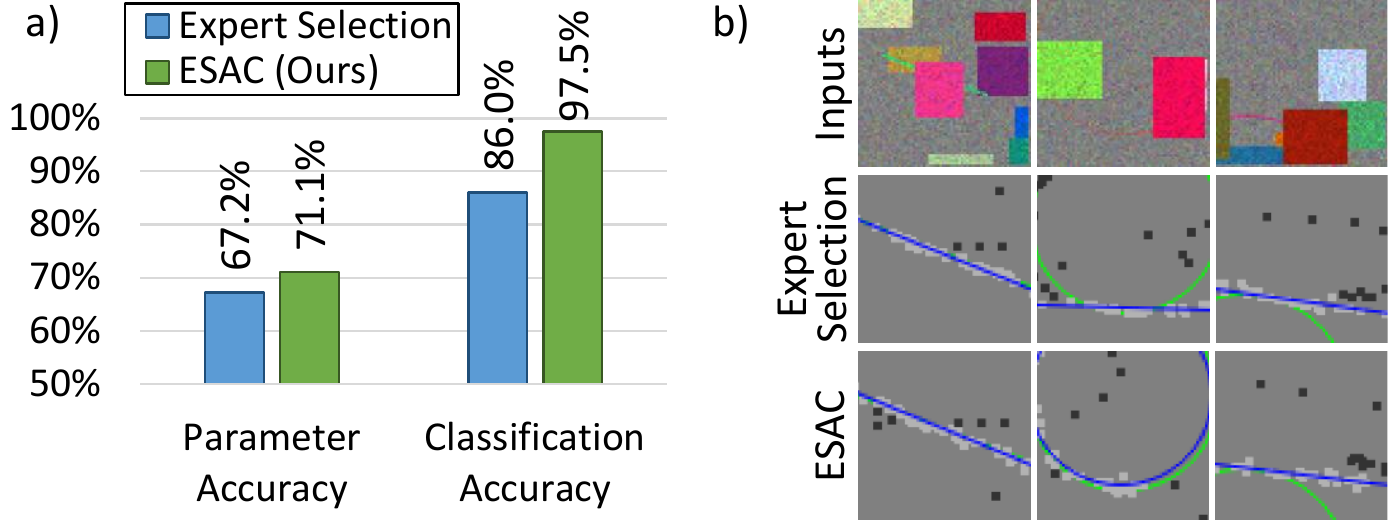}
\end{center}
\vspace{-0.3cm}
   \caption{\textbf{Results for Toy Problem. a)}
  Percentage of correctly estimated model parameters (left), and percentage of correctly selected model types, \ie line or circle (right).
 \textbf{b)} Qualitative results. The ground truth model is shown in green, the estimate is blue.}
 \vspace{-0.3cm}
\label{fig:results:toy1}
\end{figure}

\noindent{\bf Setup.}
We generate images of size \mbox{64 $\times$ 64px}, which show either a line or a circle with 50\% probability. 
We add 4 to 10 distractors to each image, which can occlude the circle or line.
Colors of lines, circles and distractors are uniformly random.
Finally, we add speckle noise to each image.
Difficult example inputs are shown in Fig.~\ref{fig:results:toy1} b).

We train one expert for lines and one for circles. 
Each expert is a CNN with 2M parameters that predicts 64 2D points.
The gating network is a CNN with 5k parameters that predicts two outputs, corresponding to the probability for a line or a circle.
As training loss for lines, we minimize the maximum distance between the estimate and ground truth in the image.
For circles, we minimize the distance between centers and absolute difference in radii of the estimate and ground truth.
We pre-train each expert using only line or only circle images with DSAC.
We pre-train the gating network using both line and circle images with a negative log likelihood classification loss.
After pre-training for 50k iterations, we train the ensemble jointly and end-to-end for another 50k iterations, either using \emph{Expert Selection} (Sec.~\ref{sec:method:moe}) or ESAC (Sec.~\ref{sec:method:esac}).
We train with a batch size of 32, using Adam \cite{adam2014}, and sampling $N=64$ model hypotheses.
For testing, we generate a set of 10,000 images.

\noindent{\bf Results.}
Fig.~\ref{fig:results:toy1} a) shows the percentage of correctly estimated model parameters (\emph{Parameter Accuracy}).
We accept a line estimate if the maximum distance to the ground truth line in the image is $<3$px.
We accept a circle estimate if its center and radius is within $3$px of ground truth.
We observe a significant advantage of using ESAC over Expert Selection (+3.9\%).
The gating network confuses images with lines and circles sometimes, and might assign higher probability to the wrong expert.
ESAC runs both experts in unclear cases, and selects the final estimate according to sample consensus.
Fig.~\ref{fig:results:toy1} a) also shows the classification accuracy of the ensemble, \ie selecting the correct model type.
Here, ESAC outperforms Expert Selection by 11.5\%.
The good classification accuracy indicates that ESAC might be a suitable method for model selection, although we did not investigate this scenario further.

\subsection{Camera Re-Localization}

For our main application, each expert predicts the same model type, a 6D camera pose, but specializes in different parts of a potentially large and repetitive environment.

\noindent{\bf Datasets.}
The \emph{7Scenes} \cite{shotton13scorf} dataset consists of RGB-D images, camera poses and 3D models of seven indoor rooms (ca.~125m$^3$ total).
The images contain texture-less surfaces, motion blur and repeating structures, which makes this dataset challenging despite its limited size. 
The \emph{12Scenes} \cite{valentin2016learning} dataset resembles 7Scenes in structure but features twelve larger rooms (ca.~520m$^3$ total).
The combination of 7Scenes and 12Scenes yields one large environment (\emph{19Scenes}) comprised of 19 rooms (ca.~645m$^3$ total, see also Fig.~\ref{fig:teaser}).
The data features multiple kitchens, living rooms and offices, containing ambiguous furniture and office equipment.

\noindent{\bf Setup.} 
Ignoring depth channels, we estimate camera poses from RGB only.
We train one expert per scene, \ie $M \in \{7,12,19\}$ depending on the dataset.
We pre-train each expert for 500k iterations, using a $L_1$ regression loss \wrt to ground truth scene coordinates obtained by rendering 3D scene models, similar to \cite{brachmann2018lessmore}.
Furthermore, we pre-train the gating network to classify scenes using negative log likelihood for 100k iterations.
We use Adam with a fixed learning rate of $10^{-4}$.
After pre-training, we train the ensemble of networks jointly and end-to-end using Expert Selection (Sec.~\ref{sec:method:moe}) or ESAC (Sec.~\ref{sec:method:esac}) for 100k iterations.
We use a learning rate of $10^{-6}$ for experts, and $10^{-7}$ for the gating network.
Otherwise, we keep the hyperparameters of DSAC++ \cite{brachmann2018lessmore}, \eg we sample $N=256$ hypotheses and use an inlier threshold of $\tau=10$px.

\begin{figure}
\begin{center}
\includegraphics[width=1\linewidth]{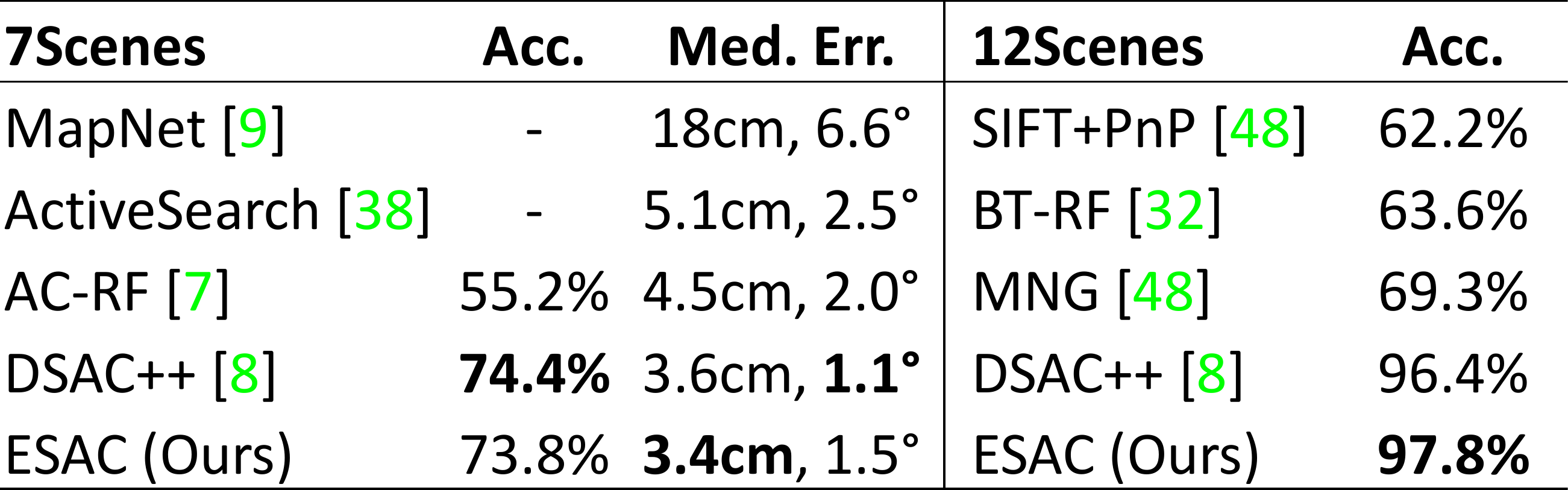}
\end{center}
\vspace{-0.3cm}
   \caption{\textbf{Pose Accuracy when Scene ID is known.} Percentage of pose estimates with an error below 5cm and 5$^\circ$, and median errors.}
\vspace{-0.5cm}
\label{fig:results:tab1}
\end{figure}

\noindent{\bf Results on Individual Scenes.}
Firstly, we verify our re-implementation of DSAC++, and our choice of network architecture.
To this end, we evaluate our expert networks when the scene ID for a test frame is given.
That is, we disable the gating network, and always use the correct expert.
We achieve an accuracy similar to DSAC++, slightly worse on 7Scenes, slightly better on 12Scenes, see Fig.~\ref{fig:results:tab1}.
Note that our networks are 7.5$\times$ smaller than those of DSAC++.

\begin{figure}
\begin{center}
\includegraphics[width=1\linewidth]{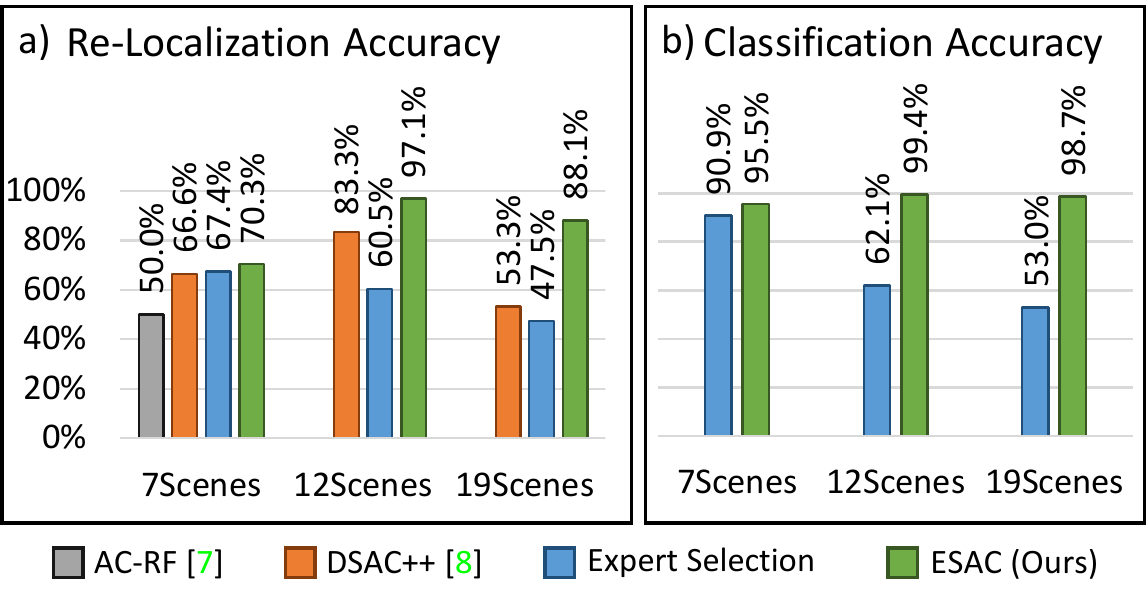}
\end{center}
\vspace{-0.3cm}
   \caption{\textbf{Average Pose Accuracy when Scene ID is Unknown. a)} Accuracy in growing environments. The scene ID has to be inferred by the method. \textbf{b)} Average accuracy of scene classification.}
 \label{fig:results:bars}
\end{figure}

\noindent{\bf Results on Combined Scenes.}
To evaluate our main contribution, we create three environments of increasing size, combining scenes of 7Scenes, 12Scenes and both (=19Scenes).
We compare to DSAC++ by training a single CNN for an environment.
For a fair comparison, we use our expert network architecture for DSAC++, and increase its capacity to match that of ESAC's network ensemble.
We also compare to an ensemble with Expert Selection (Sec.~\ref{sec:method:moe}).
We show our main results in Fig.~\ref{fig:results:bars} a) measuring the percentage of estimated poses with an error below $5^\circ$ and $5$cm.
The accuracy of DSAC++ decreases notably in larger environments, culminating in a moderate accuracy of 53.3\% re-localized images on 19Scenes.
DSAC++ relies solely on local image context which becomes increasingly ambiguous with a growing number of visually similar scenes. 
An ensemble with Expert Selection fares even worse despite using global image context in the gating network when disambiguating scenes. 
Some of the scenes are too similar, and the top-scoring gating prediction is incorrect in many cases.
By distributing model hypotheses among experts, ESAC incorporates global image context in a robust fashion, and consistently achieves best accuracy.
The margin is most distinct for 19Scenes, the largest environment, with 88.1\% correctly re-localized images.
Note that the increased environment scale hardly affects the accuracy of ESAC.
It looses 3.5\% accuracy for 7Scenes with known scene ID, and less than 1\% for 12Scenes, cf.~Fig.~\ref{fig:results:tab1}.

\begin{figure}[t!]
\begin{center}
\includegraphics[width=1\linewidth]{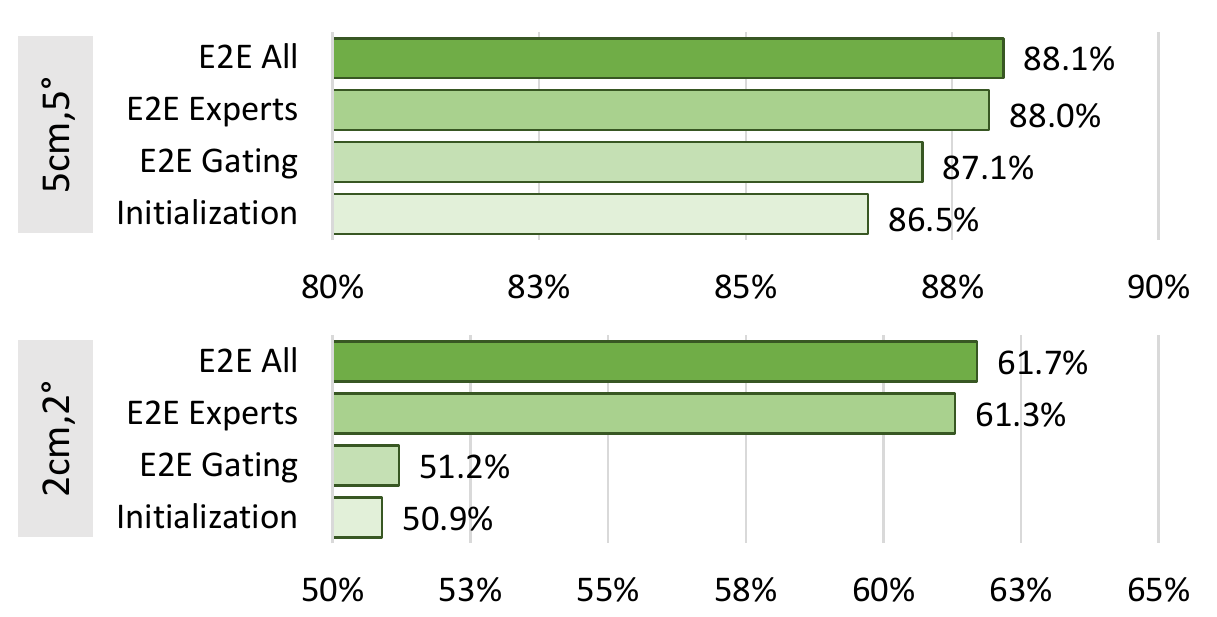}
\end{center}
\vspace{-0.2cm}
   \caption{\textbf{Effect of End-to-End Training.} Average re-localization accuracy of ESAC for 19Scenes when we train the entire ensemble of networks or parts of it end-to-end. \textbf{Top:} Acceptance threshold of 5cm and $5^\circ$. \textbf{Bottom:} 2cm and 2$^\circ$.}
   \vspace{-0.2cm}
\label{fig:ablation}
\end{figure}

\noindent{\bf Effect of End-To-End Training.}
See Fig.~\ref{fig:ablation} for the effect of end-to-end training on the overall accuracy. 
We initialize the experts by optimizing the $L_1$ distance to ground truth scene coordinates, and the gating network by optimizing the negative log likelihood classification loss (denoted \emph{Initialization}).
We then continue to train the gating network (\emph{E2E Gating}), the experts (\emph{E2E Experts}) or the entire ensemble (\emph{E2E All}) using the ESAC objective.
End-to-end training of each component increases the average accuracy, and we achieve best accuracy when we train the entire ensemble jointly.
The effect of end-to-end training is significant but not large using the common acceptance threshold of 5cm and 5$^\circ$. 
However, lowering the threshold to 2cm and 2$^\circ$ reveals a large improvement in accuracy of $>10\%$.
End-to-end training improves foremost the precision of re-localization, and less so the re-localization rate under a coarse threshold.

\noindent{\bf Handling Ambiguities.}
In Fig.~\ref{fig:results:bars} b) we show the average scene classification accuracy of Expert Selection and ESAC.
In the Appendix \ref{sec:appendix:class}, we provide additional information in the form of scene confusion matrices, and examples of visually similar scenes.
Expert Selection is particularly prone to confuse offices which contain ambiguous furniture and office equipment.
ESAC can tell these scenes apart reliably by combining global image context when distributing hypotheses and geometric consistency when selecting hypotheses. 

\begin{figure}[t!]
\begin{center}
\includegraphics[width=1\linewidth]{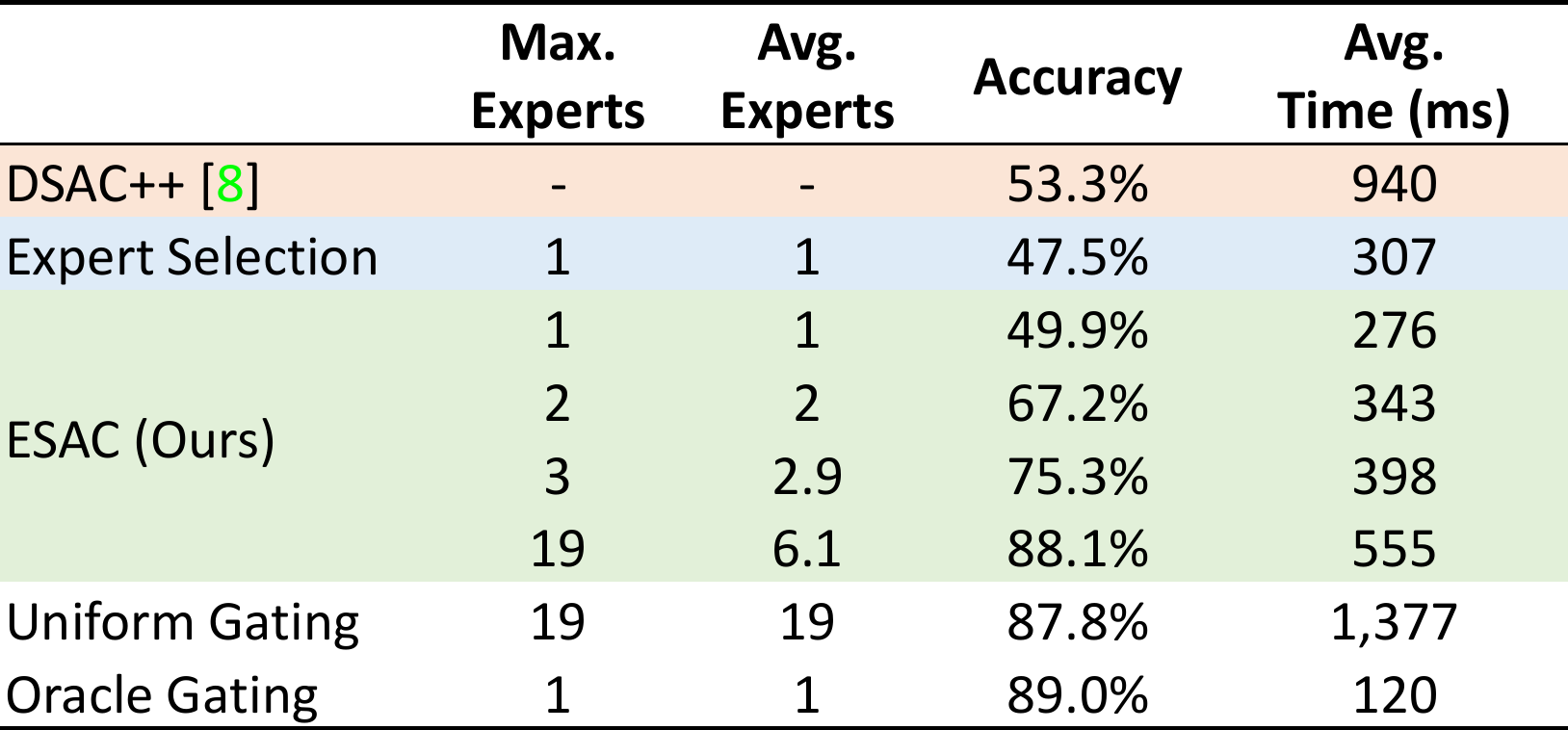}
\end{center}
\vspace{-0.3cm}
   \caption{\textbf{Accuracy vs.~Speed on 19Scenes.} We measure the average processing time for an image on a single Tesla K80 GPU including reading data. For ESAC, we can limit the maximum number of top ranked experts evaluated for a test image.}
   \vspace{0.8cm}
\label{fig:runtime}
\end{figure}

\noindent{\bf Conditional Computation.}
By using a single, monolithic network, inference with DSAC++ takes almost 1s on 19Scenes due to the large model capacity.
ESAC needs to evaluate only those experts relevant for a given test image.
On 19Scenes, it evaluates 6.1 experts in 555ms on average. 
Furthermore, we can restrict the maximum number of experts per image, see Fig.~\ref{fig:runtime}.
For example, using at most the top 2 experts per test image, we gain +19.7\% accuracy over Expert Selection with just a minor increase in computation time.
At the other end of the spectrum, we could always evaluate all experts and choose the best hypothesis according to sample consensus, see \emph{Uniform Gating} in Fig.~\ref{fig:runtime}.
This achieves good accuracy but is computationally intensive. 
ESAC shows slightly higher accuracy and is much faster.
Also, ESAC almost reaches the accuracy of \emph{Oracle Gating} which always selects the correct expert via the ground truth scene ID.

\begin{figure}
\begin{center}
\includegraphics[width=1\linewidth]{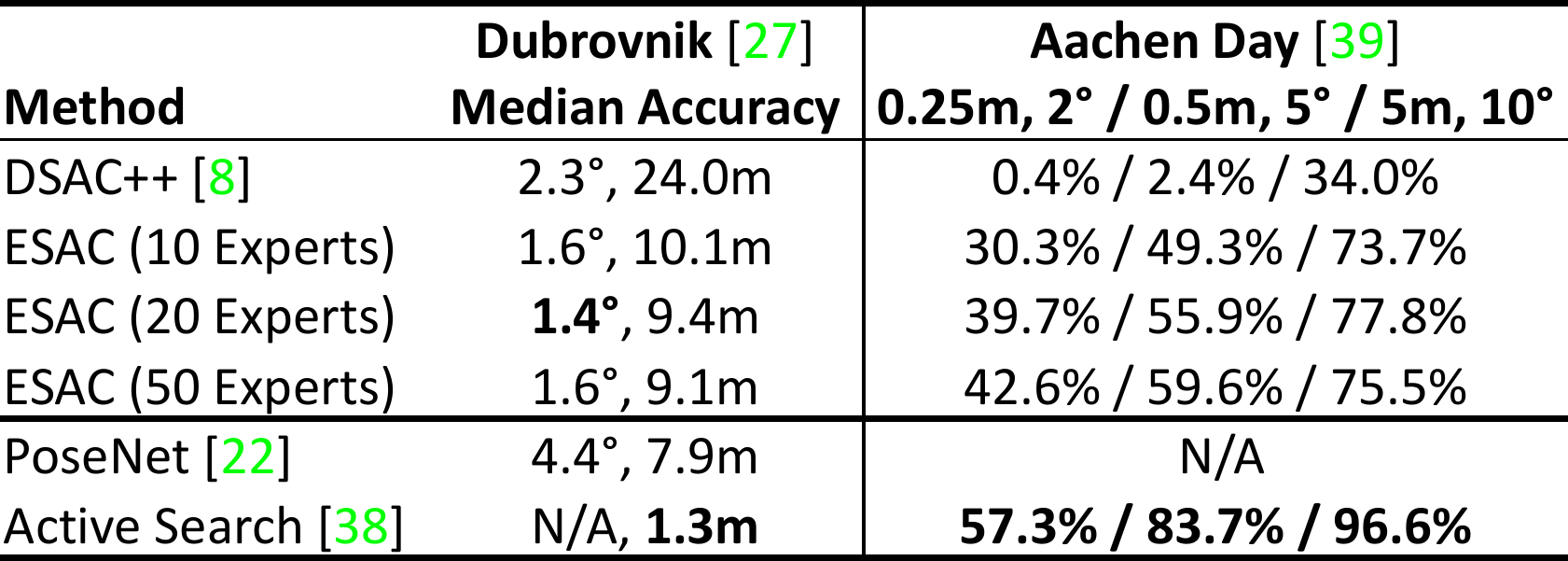}
\end{center}
\vspace{-0.15cm}
   \caption{\textbf{Large-Scale Outdoor Re-Localization.} For ESAC, we divide an environment via scene coordinate clustering, and train an expert for each cluster. See the Appendix \ref{sec:appendix:outdoor} for details.}
      %\vspace{-0.4cm}
\label{fig:results:tab2}
\end{figure}

\begin{figure}[t!]
\begin{center}
\includegraphics[width=1\linewidth]{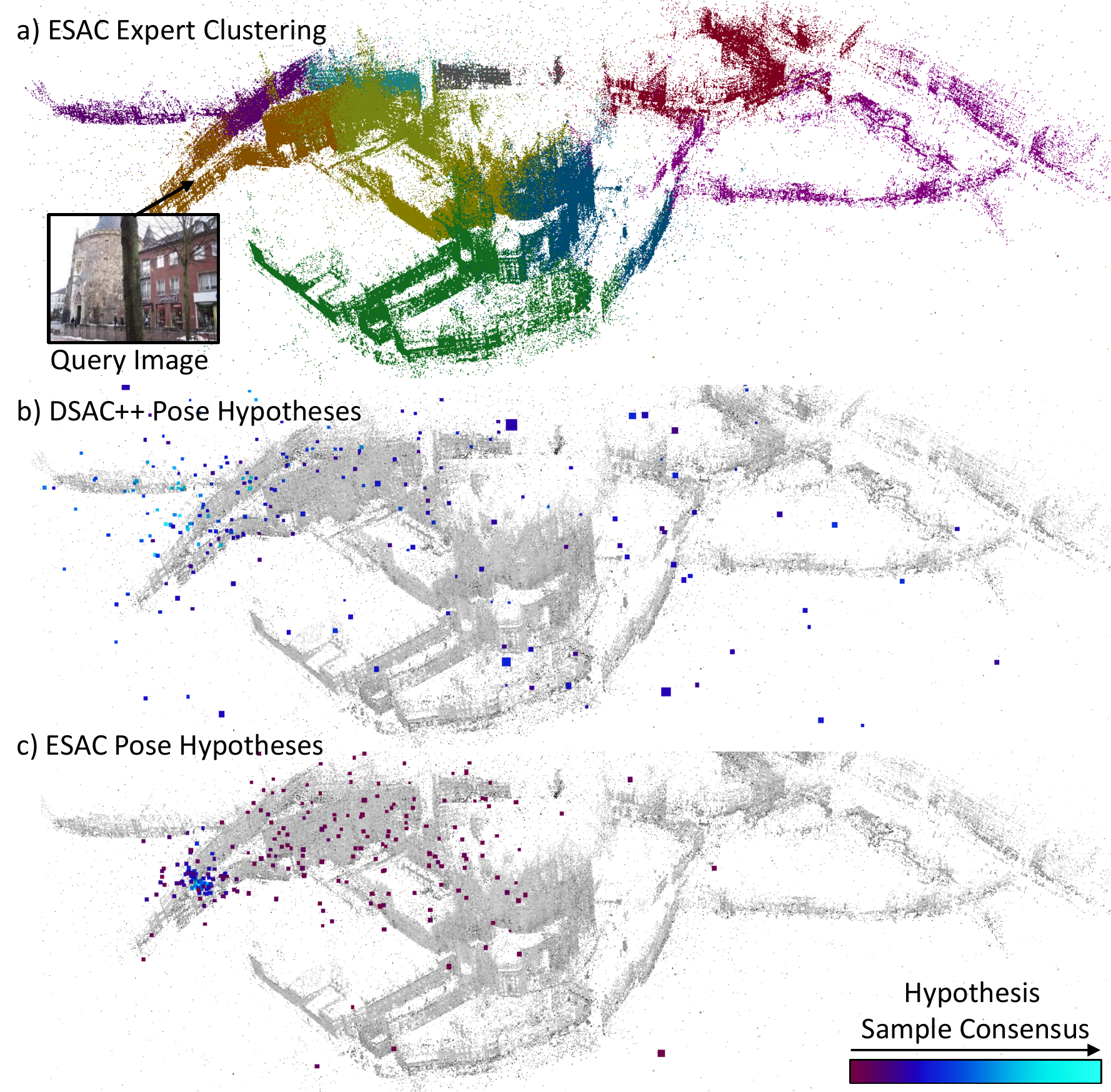}
\end{center}
\vspace{-0.1cm}
   \caption{\textbf{Qualitative Result on Aachen. a)} Clustering of the Aachen dataset used to initialize ESAC with $M=10$ experts. \textbf{b)} Positions of hypotheses drawn by DSAC++ for the query image shown in a). Hypothesis sample consensus (hypothesis score) is indicated by color, normalized over all hypotheses. \textbf{c)} Positions of hypotheses drawn by ESAC. Contrary to DSAC++, hypotheses form a clear cluster at the true position of the query image.}
   \vspace{0.13cm}
\label{fig:cluster}

\end{figure}

\noindent{\bf Outdoor Re-Localization.}
We applied ESAC to outdoor re-localization in vast connected spaces, namely to the Dubrovnik dataset \cite{li2010location}, and the Aachen Day dataset \cite{sattler2018changing}.
Appendix \ref{sec:appendix:outdoor} contains details about the experimental setup. 
We present the main results in Fig.~\ref{fig:results:tab2} and a qualitative result in Fig.~\ref{fig:cluster}.
While we improve over DSAC++ by a large margin, we do not completely close the performance gap to classical sparse feature-based methods like ActiveSearch \cite{sattler2016efficient}.
Adding more experts (and therefore model capacity) helps only to some degree.
This hints towards limitations of current scene coordinate regression methods \cite{brachmann2017dsac, brachmann2018lessmore} beyond the environment size.
For example, the SfM ground truth reconstruction, which we use for training, contains a substantial amount of outliers, particularly for Dubrovnik, see Appendix \ref{sec:appendix:outdoor} for a detailed discussion.
The training of CNN-based dense regression might be sensitive to such noisy inputs, and developing resilient training strategies might be a promising direction for future research.

\vspace{-0.2cm}
\section{Conclusion}
\label{sec:discussion}
\vspace{-0.1cm}

We have presented ESAC, an ensemble of expert networks for estimating parametric models. 
ESAC uses a gating network to distribute model hypotheses among experts.
This is more robust than formulations where the gating network chooses a single expert only.
We applied ESAC to the camera re-localization task in a large indoor environment where each expert specializes to a single room, achieving state-of-the-art accuracy.
For large-scale outdoor re-localization, we made progress towards closing the gap to classical, feature-based methods.

\nocite{meng17}

\vspace{-0.3cm}
\paragraph*{Acknowledgements:}
This project has received funding from the European Research Council (ERC) under the European Union’s Horizon 2020 research and innovation programme (grant agreement No 647769). The computations were performed on an HPC Cluster at the Center for Information Services and High Performance Computing (ZIH) at TU Dresden.

\appendix
\vspace{0.5cm}
\section{Scene Classification}
\label{sec:appendix:class}

\begin{figure*}
\begin{center}
\includegraphics[width=1.0\linewidth]{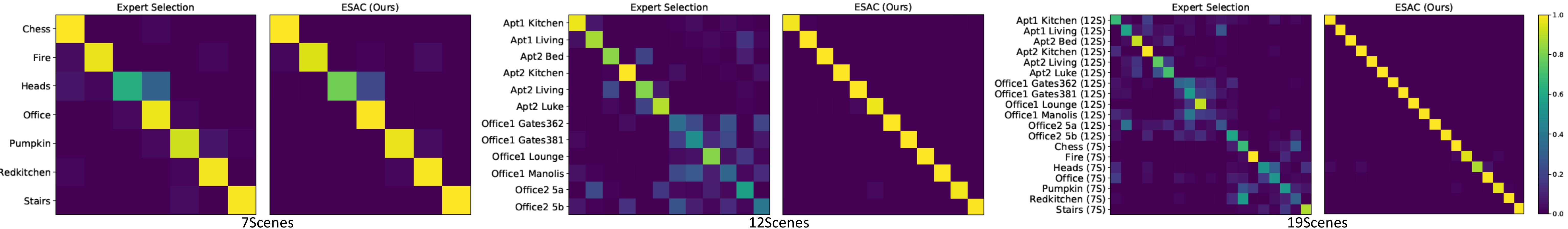}
\end{center}
\vspace{-0.2cm}
   \caption{\textbf{Scene Confusion.} We compare confusion matrices of Expert Selection and ESAC for 7Scenes, 12Scenes and 19Scenes. The Y-axis shows the true scene, the X-axis shows the estimated scene. The scene ordering on the X-axis (from left to right) follows the Y-axis (from top to bottom). For 19Scenes, we mark scenes originating from 12Scenes with \emph{12S} and scenes originating from 7Scenes with \emph{7S}.}
\label{fig:confusion}
\end{figure*}

\begin{figure*}
\begin{center}
\includegraphics[width=0.9\linewidth]{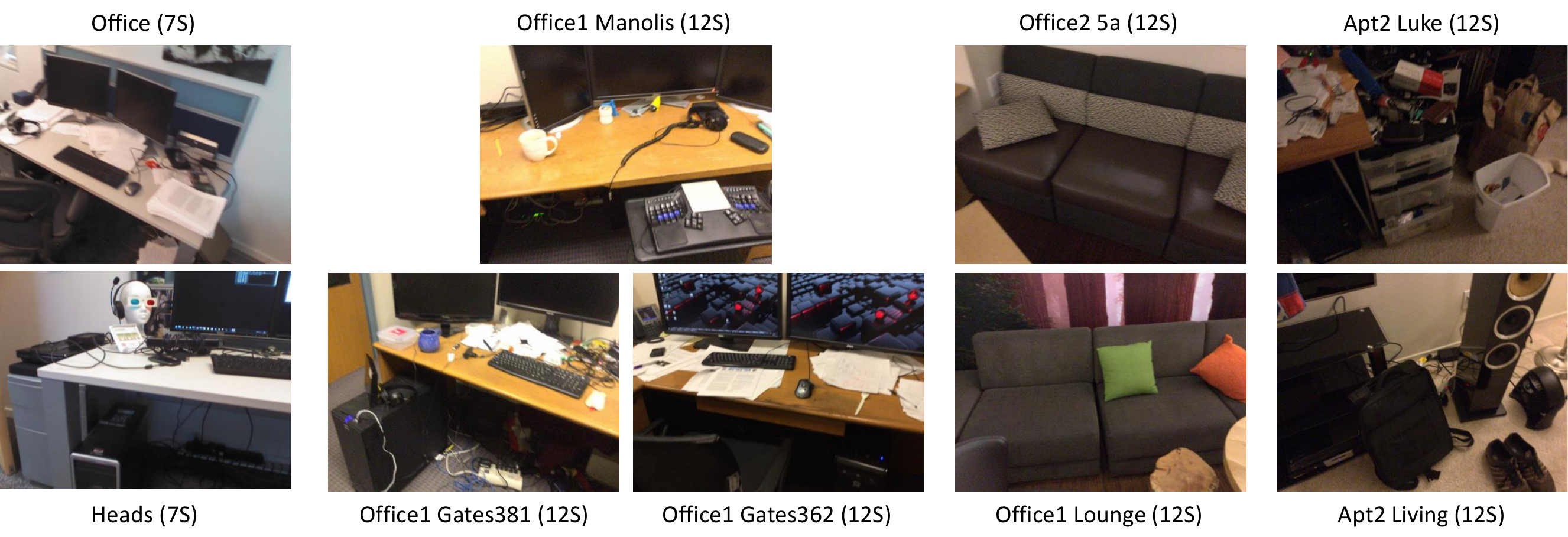}
\end{center}
\vspace{-0.3cm}
   \caption{\textbf{Ambiguous Scenes.} We show test frames of scenes that Expert Selection confuses often (cf.~Fig.~\ref{fig:confusion}). We mark scenes originating from 12Scenes with \emph{12S} and scenes originating from 7Scenes with \emph{7S}.}
   %\vspace{-0.3cm}
\label{fig:confusion2}
\end{figure*}

In Fig.~\ref{fig:confusion}, we show the scene confusion matrices of our ensemble trained with Expert Selection and ESAC.
The database contains multiple offices which look similar due to ambiguous office equipment, see Fig.~\ref{fig:confusion2} for examples.
Expert Selection chooses a scene according to the prediction of the gating network, which is error prone.
ESAC considers multiple experts in ambiguous cases, and chooses the final estimate according to geometric consistency.

\section{Large-Scale Outdoor Re-Localization}
\label{sec:appendix:outdoor}

We provide details regarding our experiments on large-scale outdoor re-localization. 
We apply ESAC to two large outdoor re-localization datasets, namely Dubrovnik \cite{li2010location} and Aachen Day \cite{sattler2018changing}.

\noindent \textbf{Datasets.}
The Dubrovnik dataset is comprised of ca.~6k holiday photographs taken in the old town of Dubrovnik. 
Stemming from online photo collections, the images were recorded with different cameras, and feature a multitude of different focal lengths, resolutions and aspect ratios.
The Aachen Day dataset is comprised of ca.~4.5k images taken in Aachen, Germany. 
Training and test images were recorded using two separate but comparable camera types.
The full Aachen dataset also comes with a small collection of difficult night time query images (\emph{Aachen Night}) which we omit here.
There is no night time training data, and bridging the resulting domain gap is out of scope of ESAC.

\noindent \textbf{ESAC Training.}
Both datasets represent large connected areas.
For initializing the gating network and scene coordinate experts, we divide each area into clusters via kMeans, see Fig.~\ref{fig:cluster} a) for an example. 
As input for clustering, we use the median scene coordinate (median per dimension) for each training image.
To avoid quantization effects at the cluster borders during initialization, we use the following soft assignment of training images to experts.
We express the probability $p_e(I)$ of training image $I$ belonging to the cluster of expert $e$ via a similarity measure $S(I, e)$:
\begin{equation}
p_e(I) = \frac{S(I, e)}{\sum_{e'} S(I, {e'})}.
\end{equation}
We define this similarity in terms of the distance between the mean scene coordinate of image $I$, denoted $\bar{\crd}_I$, and the cluster center $\mathbf{c}_e$:
\begin{equation}
S(I,e) = \frac{1}{2\pi\sigma_e}\exp{\frac{-\gamma||\bar{\crd}_I - \mathbf{c}_e||_2}{2\sigma_e}},
\end{equation}
where $\sigma_e$ is an estimate of the cluster size, and $\gamma$ controls the softness of the similarity.
We use $\gamma = 5$, and the mean squared distance of all images (resp.~their median scene coordinates) within a cluster to the cluster center as $\sigma_e$.
When initializing the gating network, me minimize the KL-divergence of gating predictions $g(e, I, \param)$ and probabilities $p_e(I)$.
When initializing an expert network $e$, we randomly choose training images according to $p_e(I)$ and minimize the $L_1$ distance \wrt ground truth scene coordinates for 1M iterations.
We obtain ground truth scene coordinates by rendering the sparse SfM reconstruction using the ground truth pose for image $I$.
Since ground truth scene coordinates are sparse, we optimize the re-projection error of the dense scene coordinate prediction for another 1M iterations, hence following the two-stage initialization of DSAC++ \cite{brachmann2018lessmore}.
Finally, we train the entire ensemble jointly and end-to-end for 50k iterations using the ESAC objective. 
To support generalization to different camera types and lighting conditions, we convert all images to grayscale, and randomly change brightness and contrast during training in the range of 50-110\% and 80-120\%, respectively.

\noindent \textbf{Discussion of Results.}
As stated in the main text, ESAC demonstrates largely improved accuracy on both outdoor datasets compared to DSAC++ \cite{brachmann2018lessmore}. 
However, is does not yet reach the accuracy of  ActiveSearch \cite{sattler2016efficient}, a classic sparse feature-based re-localization method.
Especially on Dubrovnik, ESAC stays far behind, even when using a substantial amount of $M=50$ experts.

Upon closer inspection, we find that the structure of these datasets potentially contributes to the exceptional performance of ActiveSearch.
Both datasets come with a 3D model of the environment and ground truth training poses created by running a sparse feature-based structure-from-motion reconstruction tool on all images (training and test).
Images which are challenging for feature-based approaches (\ie images with little structure or motion blur) are naturally not part of these datasets, since they are filtered at the reconstruction stage.
It might be problematic to compare learning-based approaches to classical feature-based methods on datasets, where the ground truth was generated with feature-based reconstruction tools.

\begin{figure}[t!]
\begin{center}
\includegraphics[width=1\linewidth]{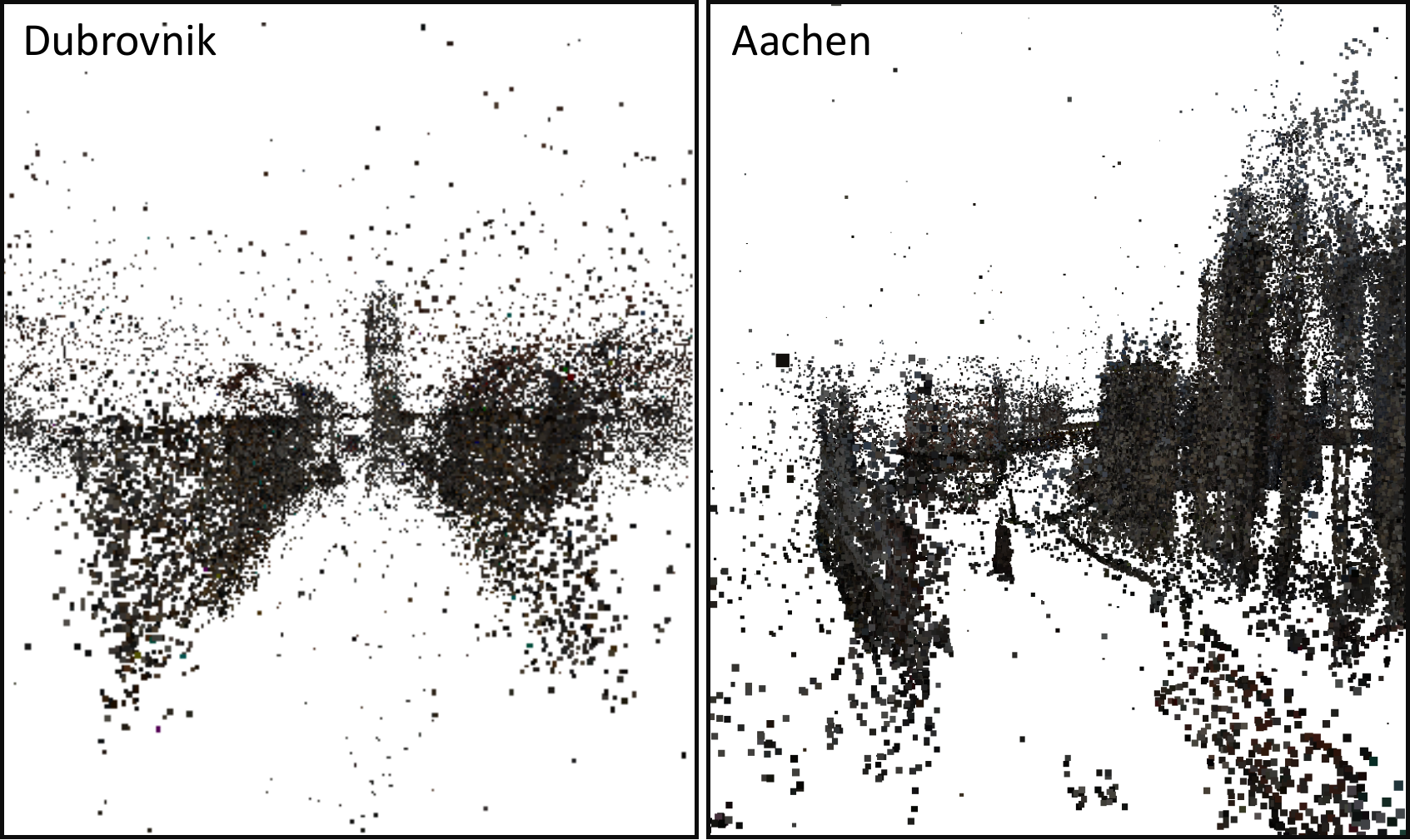}
\end{center}

   \caption{\textbf{Ground Truth SfM Reconstructions.} Both datasets contain outlier 3D points in the ground truth reconstruction. The outlier ratio is substantial for Dubrovnik, and still noticeable for Aachen.}
   \vspace{0.6cm}
\label{fig:outliers}
\end{figure}

Furthermore, the reconstructions are not perfect as they contain a substantial amount of outlier points, see Fig. ~\ref{fig:outliers} for an illustration.
ActiveSearch operates directly on top of this reconstruction, and applies sophisticated outlier rejection schemas.
In contrast, scene coordinate regression methods like ESAC try to build geometrically consistent internal representations of a map, encoded in the network weights.
Having visually similar image patches associated with very different ground truth scene coordinates (due to outliers) might result in severe overfitting of the network, which tries to tell patches apart that actually show the same location.
The poor accuracy of ESAC on Dubrovnik compared to Aachen supports this interpretation, as the re-localization accuracy corresponds well to the general reconstruction quality of both datasets.
At the same time, the question arises how meaningful the reported $\approx$1m re-localization accuracy for ActiveSearch on the Dubrovnik dataset is, given the ground truth quality.
Note that geometry, training poses and test poses were all jointly optimized during the SfM reconstruction. 
Inaccuracies in the geometry might therefore hint towards limited accuracy of the ground truth poses.

{\small
\bibliographystyle{ieee_fullname}
\bibliography{esac}
}
\end{document}